\documentclass{IOS-Book-Article}%[seceqn,secfloat,secthm]

\usepackage[utf8]{inputenc}
\usepackage[pdftex,bookmarks,colorlinks=false]{hyperref}
\usepackage{graphicx}
\usepackage{filecontents}
\usepackage{setspace}
\usepackage{amsmath}
\usepackage{xspace}
\newcommand{\footnoteurl}[1]{{\tiny \url{#1}}}
\newcommand{\areaname}[1]{{\bf #1}}

\begin{document}
\begin{frontmatter}          % The preamble begins here.

 \title{Towards Meaningful Maps of Polish Case~Law}

\runningtitle{Towards Meaningful Maps of Polish Case Law}

% Two or more authors:
\author[A,B]{\fnms{Michał} \snm{Jungiewicz}},
\author[A]{\fnms{Michał} \snm{Łopuszyński}}
\runningauthor{M. Jungiewicz and M. Łopuszyński}
\address[A]{Interdisciplinary Centre for Mathematical and Computational
Modelling, University of Warsaw, Pawińskiego 5a, 02-106 Warsaw, Poland}
\runningauthor{M. Łopuszyński}
\address[B]{Faculty of Electronics and Information Technology,
 Warsaw University of Technology, Nowowiejska 15/19, 00-665 Warsaw, Poland}
\end{frontmatter}

{\scriptsize
This a draft version of the paper published in ``Legal Knowledge an
Information Systems, JURIX 2015: The Twenty-Seventh Annual Conference'',
Frontiers in Artificial Intelligence and Applications, Volume 279, edited
by Antonino Rotolo, IOSPress, 2015.  The final publication is available
from \href{http://dx.doi.org/10.3233/978-1-61499-609-5-185}{IOSPress}}.
\smallskip

In this work, we analyze the utility of two dimensional document maps for
exploratory analysis of Polish case law. Such maps reflect the
structure of analyzed collection by grouping similar documents in a
neighbouring regions of 2D space. This visual aid could be useful for
browsing and searching, finding anomalous documents or
quickly gaining synthetic knowledge about large corpora.

We started our study by selecting two automatic methods of generating document maps.
First one is a standard method using linear principal component analysis (PCA) and
visualizing two most important components on a scatter-plot. Second one is a
modern nonlinear method of embedding multidimensional data in~2D, namely, the
t-Distributed Stochastic Neighbor Embedding method (t-SNE)
\cite{vanDerMaaten2008}.  t-SNE is optimized to preserve local structure
(i.e., very similar documents should end up close together) and to reduce
the so called ,,crowding problem''.  The ,,crowding problem'' is undesired
feature of many dimensionality reduction techniques, which essentially place
many data points in a very small area of low-dimensional space (hence
creating a crowd). These two features make t-SNE unique and frequently
used in recent analyses of high-dimensional data.

\begin{figure}[!h]
\centering
\includegraphics[width=0.30\textwidth]{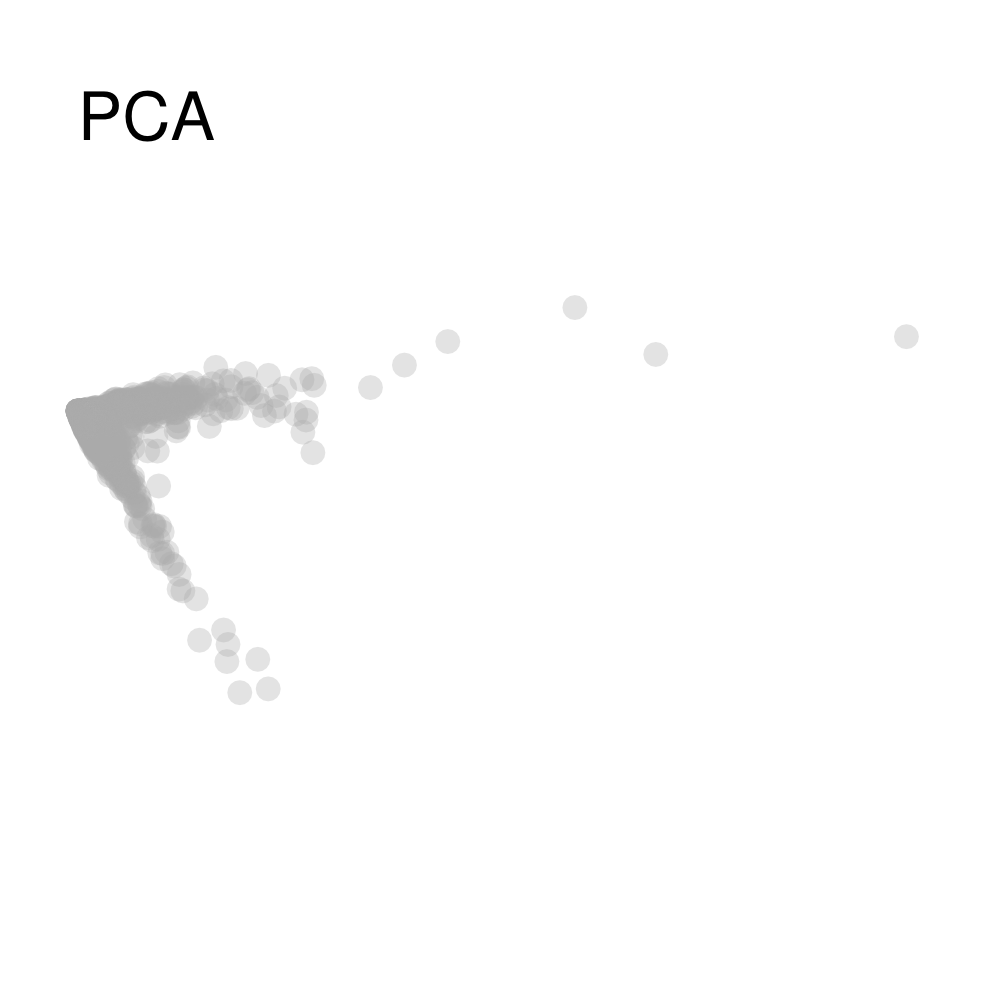}
\includegraphics[width=0.30\textwidth]{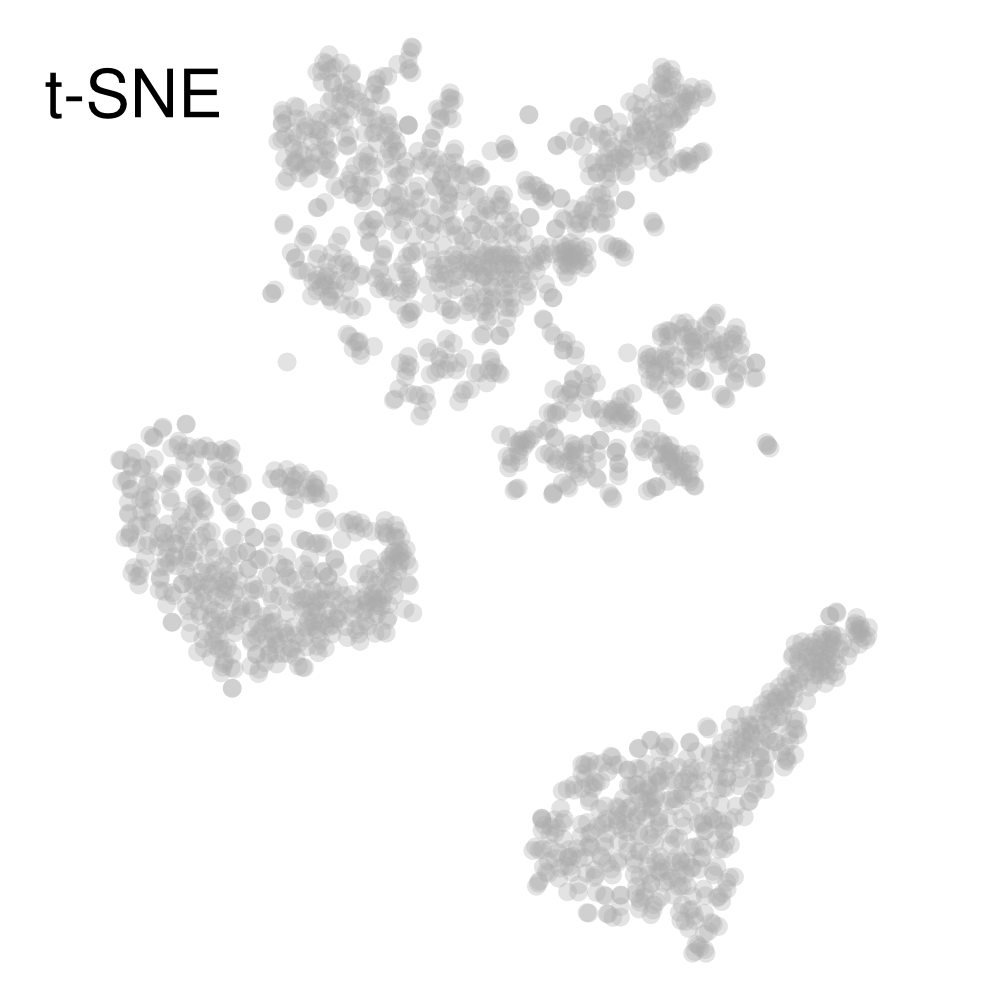}
\vspace{-0.4cm}
\caption{Document maps created using PCA and t-SNE methods from 2000 judgments
         from Supreme Court (500 docs.), Constitutional Tribunal (500 docs.),
         common courts (500 docs.), and National Appeal Chamber (500 docs.)
         \label{fig:tSNEandPCAmono}}
\end{figure}

\begin{figure}[!h]
\centering
\includegraphics[width=0.75\textwidth]{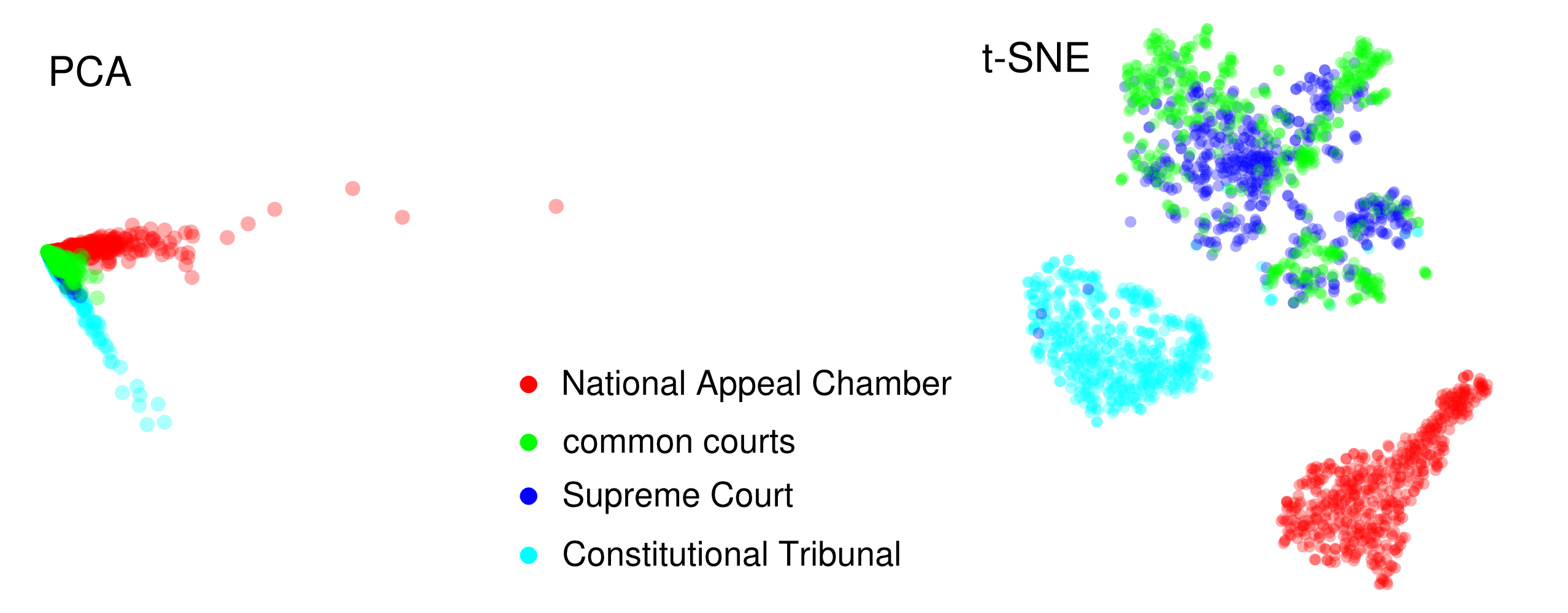}
\vspace{-0.3cm}
\caption{Map from Figure~\ref{fig:tSNEandPCAmono} with added coloring related to
         issuing institution \label{fig:tSNEandPCAcolour}}
\end{figure}

The results of the PCA and t-SNE analysis of 2000 judgments from Polish courts
are presented in Figure~\ref{fig:tSNEandPCAmono}.
t-SNE approach forms three well separated clusters which occupy large
fraction of available space. For PCA, there is no obvious structure visible,
as well as most of the available space remains unused.  In
Figure~\ref{fig:tSNEandPCAcolour}, we added color-coded information about
the document source (which was not directly used as a feature for computing
the mapping!) to Figure~\ref{fig:tSNEandPCAmono}. We find that indeed
clusters visualized by t-SNE correspond to the judicial bodies. The common
court judgments were merged together with the Supreme Court documents by \mbox{t-SNE},
hence three not four groups. In Poland, the Supreme Court is the court of last
resort of appeal against judgments of common courts, so this artifact can be
easily justified.

\begin{figure}[!ht]
\centering
\includegraphics[width=0.35\textwidth]{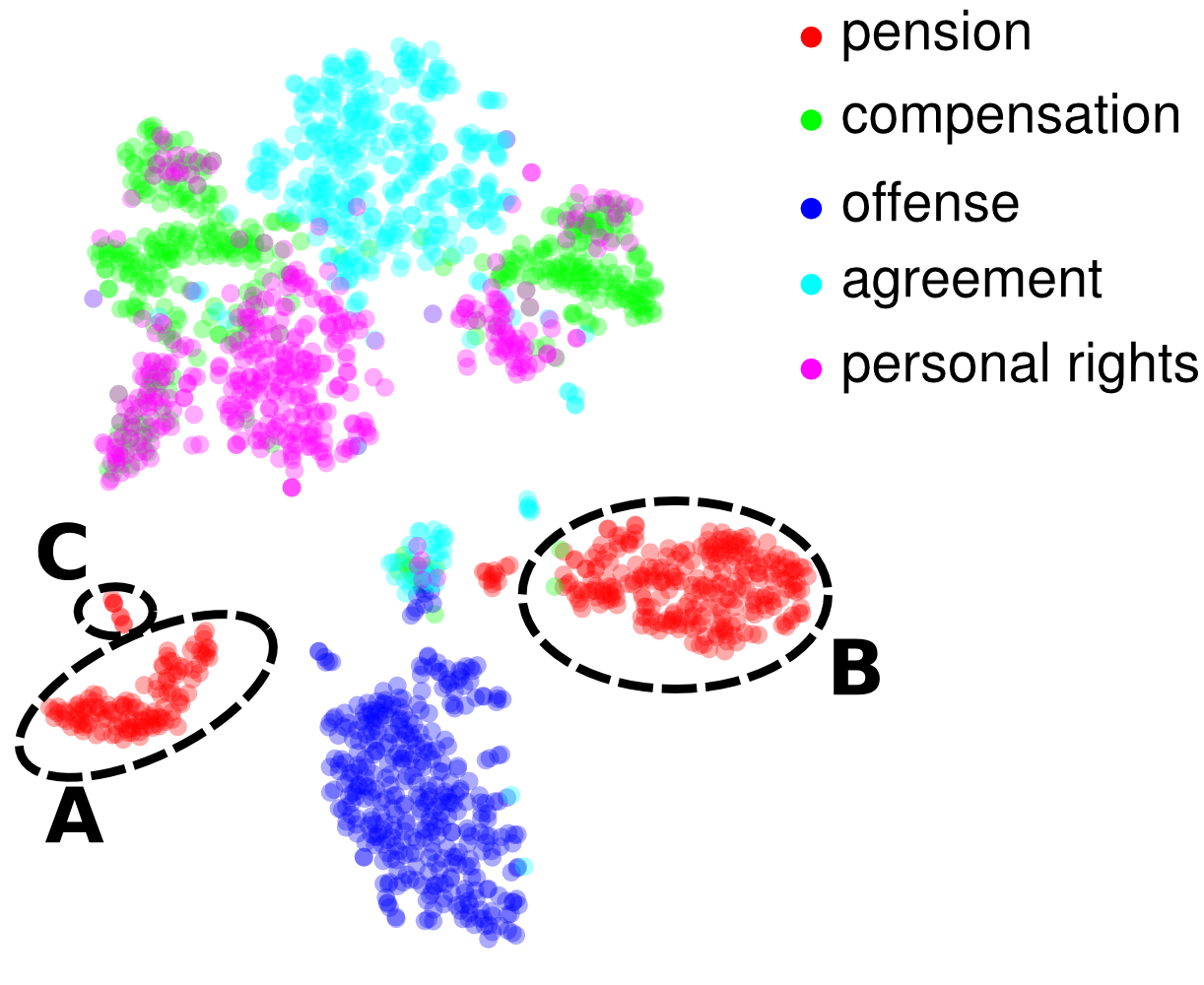}
\vspace{-0.4cm}
\caption{t-SNE visualization of 2000 judgements from common courts in Poland
         tagged by different keywords (400 judgments per each keyword)
         \label{fig:tSNEKeywords}}
\end{figure}

To further explore the utility of the t-SNE method, we visualized
2000 common court judgments selected randomly from documents tagged by
five different keywords, see Figure~\ref{fig:tSNEKeywords}.
The graph generated by t-SNE mostly groups together documents labeled by the
same keyword. Manual analysis of separated mapping regions related to
word ,,pension'' reveals hidden topical structure in the judgments.  Region
\areaname{A} turns out to correspond to cases about increase/recalculation
of pension, \areaname{B} deals with actually granting a pension,
\areaname{C} contains cases related to military pensions for which special
regulations exist in Poland. This shows the potential of the t-SNE method
in the exploratory analysis of judgments corpora.

To summarize, in our opinion t-SNE seems a promising approach to exploratory
analysis of case law. After further investigation, t-SNE document maps
may prove to be useful addition to legal information systems.

{\small
{\bf Acknowledgments.} We acknowledge the support from the SAOS
(\href{http://saos.org.pl}{http://saos.org.pl}) project financed by the
National Centre for Research and Development.}
%%%%%%%%%%% The bibliography starts:
\bibliographystyle{unsrt}
\bibliography{biblio}
\end{document}